\documentclass{article}

\PassOptionsToPackage{numbers,sort&compress}{natbib}
\usepackage[preprint]{arxiv_style}

\usepackage[utf8]{inputenc}
\usepackage[T1]{fontenc}
\usepackage[colorlinks=true,linkcolor=red!60!black,citecolor=blue!60!black,urlcolor=teal!70!black]{hyperref}
\usepackage{url}
\usepackage{times}
\usepackage{latexsym}
\usepackage{microtype}
\usepackage{inconsolata}
\usepackage{graphicx}
\usepackage{booktabs}
\usepackage{amsfonts,amsmath,amssymb}
\usepackage{nicefrac}
\usepackage{xcolor}
\usepackage{multirow}
\usepackage{xspace}
\usepackage{array}
\usepackage{algorithm}
\usepackage{algpseudocode}

\newcommand{\ours}{\textsc{RCDA}\xspace}

\title{Recipe-Controlled Decoder Audit for Structural Knowledge-Graph Completion%
  \thanks{Code and artifacts are available at \href{https://github.com/AndyShan11/kgc-decoder-audit}{github.com/AndyShan11/kgc-decoder-audit}.}}

\author{%
  Xihang Shan \\
  School of Mathematical Sciences \\
  Xiamen University \\
  Fujian, China \\
  \texttt{19020232202354@stu.xmu.edu.cn} \\
  \And
  Ye Luo\thanks{Corresponding author.} \\
  School of Informatics \\
  Xiamen University \\
  Fujian, China \\
  \texttt{luoye@xmu.edu.cn} \\
}

\begin{document}

\maketitle
\begin{abstract}
We present a recipe-controlled decoder audit (\ours) for structural transductive knowledge-graph completion (KGC). The audit asks a simple reporting question: before attributing gains to an encoder or training recipe, what changes when the decoder is swapped under the same recipe? Using ComplEx and DistMult as the primary controlled pair, with targeted RotatE/TransE spot-checks, we evaluate seven benchmarks. On five standard KGs, ComplEx-vs-DistMult differences are modest but consistent under our recipe ($+0.005$ to $+0.012$ MRR), whereas CompGCN-style encoder effects vary more by dataset. On small KGs, decoder effects become the main diagnostic: Kinship shows a stable ComplEx advantage of $+0.143$ MRR (6 seeds), while UMLS favours ComplEx by $+0.022$ MRR in a clean 6-seed server rerun but reverses in an earlier provenance variant. We therefore treat small-KG decoder choice as recipe- and provenance-sensitive rather than as a fixed dataset winner. We further show that decoder choice interacts with encoder depth on WN18RR, and that under our recipe $L{=}0$ ComplEx on YAGO3-10 reaches $0.6971 \pm 0.0048$ MRR at $d{=}128$. The result is a compact audit protocol: report matched decoder rows, log small-KG provenance, and sweep decoder $\times$ depth before making encoder-level claims.

\end{abstract}

\section{Introduction}
\label{sec:intro}

A structural KG-completion model has three independent levers: the \emph{decoder} (ComplEx~\cite{complex}, DistMult~\cite{distmult}, RotatE~\cite{rotate}, $\ldots$), the \emph{encoder} (e.g., CompGCN~\cite{compgcn} message passing), and the \emph{training recipe}. Prior audits cover recipe choice~\cite{ruffinelli2020olderboys} and CompGCN-style encoders~\cite{zhang2022rethinking}; the head-to-head DistMult-vs-ComplEx decoder axis under one recipe has not been audited at comparable breadth. We study structural transductive KGC on seven benchmarks, primarily with ComplEx/DistMult and CompGCN-style encoders, plus a small RotatE/TransE spot-check.

\paragraph{Headline: decoder choice is a reporting confound.}
Our goal is not to introduce a new KGC architecture or to rank decoders universally. Instead, we turn decoder choice into an explicit audit axis. On standard large KGs, the ComplEx-vs-DistMult gap is small but consistent under our recipe; on small KGs, the same swap can change the diagnostic conclusion; and on WN18RR, the best encoder depth depends on the decoder. The actionable recommendation is therefore narrow and practical: structural KGC papers should report a matched DistMult-vs-ComplEx row, and should avoid treating decoder choice as an implementation detail when making encoder or recipe claims.

\paragraph{Contributions.}
(C1) We propose \ours, a recipe-controlled decoder-audit protocol for structural KGC. (C2) We provide a seven-benchmark ComplEx/DistMult diagnostic with targeted RotatE/TransE spot-checks, separating shared-grid evidence from small-KG decoder-only sensitivity. (C3) We show that Kinship has a large stable ComplEx advantage, while UMLS changes winner across provenance variants; edges/relation and symmetry are useful descriptors but not winner rules. (C4) We show decoder--encoder interaction on WN18RR: optimal $L$ depends on the decoder. (C5) We provide a recipe-controlled structural reference grid, including $0.6971 \pm 0.0048$ MRR on YAGO3-10 ($d{=}128$) and $0.5379$ on CoDEx-M.

\section{Related Work}
\label{sec:related}

\paragraph{Prior audits.}
Recipe audits show that older KGE models can match newer methods under careful training~\cite{ruffinelli2020olderboys,ali2021hparam,akrami2020realistic,sun2020reevaluation}. Encoder audits decompose R-GCN/CompGCN-style gains and find that graph-structure modelling is often less decisive than expected~\cite{zhang2022rethinking}. Our increment is orthogonal: we hold one recipe fixed, vary the decoder, and report when the decoder check changes the diagnostic conclusion.

\paragraph{The decoder axis: prior conceptual hypothesis, no direct measurement.}
The decoder axis has been studied at the level of single-dataset benchmark numbers but not as a controlled comparison. The closest prior is \cite{manabe2018datadependent} (AAAI 2018), who proposed an L1 regulariser on ComplEx imaginary parts that pushes them to zero for empirically-symmetric relations, on the conceptual ground that ``parameters in [ComplEx] are superfluous for relations that are either symmetric or antisymmetric.'' They did not measure how a vanilla DistMult-vs-ComplEx choice behaves under matched structural recipes. We do (Sec.~\ref{sec:umls-spotlight}); we also report that low-edge small KGs are recipe-sensitive, so relation symmetry and edges/relation should be treated as audit descriptors rather than sufficient winner rules. \cite{kazemi2018simple} introduced SimplE which interpolates between DistMult and ComplEx via two embedding heads. \cite{trouillon2017complex} (the JMLR-extended ComplEx paper) reports DistMult and ComplEx side-by-side on FB15k and WN18 but does not include the low-edges-per-relation benchmarks (UMLS, Kinship) where the decoder axis is most sensitive.

\paragraph{Decoder and method families.}
Shallow embedding models such as TransE~\cite{transe}, DistMult~\cite{distmult}, ComplEx~\cite{complex}, RotatE~\cite{rotate}, TuckER~\cite{balazevic2019tucker}, and HAKE~\cite{hake} score triples from endpoint embeddings. GNN-encoder extensions add message passing~\cite{rgcn,compgcn,chen2021hitter}; path-based reasoners such as NBFNet, A$^*$Net, and RED-GNN encode query-specific structure~\cite{nbfnet,astarnet,redgnn}. The closest decoder-axis hypothesis is \cite{manabe2018datadependent}, which regularises ComplEx imaginary parts for empirically symmetric relations; however, it does not audit the vanilla decoder choice under matched recipes. We also revisit the YAGO3-10 ComplEx-N3 line~\cite{lacroix2018complex}: under our recipe, $L{=}0$ ComplEx is already near saturation at $d{\approx}128$, with only a small single-seed gain at $d{=}256$.

\section{Approach: Recipe-Controlled Decoder Audit}
\label{sec:method}

The audit varies three levers under one fixed training recipe: (i) \textbf{primary decoder} $\in \{$ComplEx, DistMult$\}$, with targeted RotatE/TransE checks; (ii) \textbf{encoder depth} $L \in \{0, 2, 3\}$ ($L{=}0$ = pure embedding lookup); and (iii) \textbf{recipe} ablations. We refer to the protocol as \ours\ (\textbf{R}ecipe-\textbf{C}ontrolled \textbf{D}ecoder \textbf{A}udit).

\paragraph{Encoder.}
A standard CompGCN~\cite{compgcn} layer with Hadamard composition aggregates relation-conditioned neighbour messages, with residuals and layer norm. We use $L \in \{2,3\}$ unless scanning depth, hidden $d{=}200$ on FB/WN and $d \in \{64,128\}$ on YAGO3-10.

\paragraph{Decoders.}
DistMult scores $\langle \mathbf{h}_h \odot \mathbf{r}, \mathbf{h}_t \rangle$. ComplEx splits embeddings into real/imaginary halves and can model antisymmetric relations. We compare the common-practice equal-$d$ setting rather than a parameter-matched setting; this is intentional because the audit targets the reporting choices typically hidden in structural KGC papers.

\paragraph{Training recipe.}
\label{sec:recipe}
We train with 1-vs-all CE over all entities, label smoothing $\varepsilon{=}0.3$, AdamW, MultiStepLR drops at 60\% and 80\%, dropout $0.2$, batch $1024$/$2048$, and weight decay $10^{-4}$. Training runs for 300/200/150 epochs on WN/FB/YAGO-style scales.

\section{Experiments}
\label{sec:experiments}

\subsection{Setup}
\label{sec:setup}

\paragraph{Datasets and evaluation.}
We evaluate five standard transductive benchmarks (FB15k-237~\cite{fb15k237}, WN18RR~\cite{wn18rr}, YAGO3-10~\cite{yago310}, CoDEx-M, CoDEx-L~\cite{safavi2020codex}) plus UMLS and Kinship for the small-KG decoder spotlight. We report filtered MRR / Hits@$\{1,3,10\}$ under the standard protocol~\cite{transe} with average-rank tie-breaking. CoDEx-M/CoDEx-L use the canonical Safavi et al.\ splits; inverse triples are added only to training. Headline rows use 6 seeds where stated and otherwise 3 seeds; the exact seed lists are retained in the run JSONs/logs.

\paragraph{Recipe and reproducibility details.}
Defaults are $d{=}200$ for FB, WN, CoDEx, UMLS, and Kinship; $d\in\{64,128\}$ for YAGO3-10; $L\in\{0,2,3\}$; batch $1024$ or $2048$ (CoDEx-L $4096$); dropout $0.2$; AdamW with lr $5{\times}10^{-4}$ and weight decay $10^{-4}$; MultiStepLR at $0.6T/0.8T$ with factor $0.3$; label smoothing $\varepsilon{=}0.3$; and 1-vs-all CE over all entities. Training uses AMP/FP16 on a single NVIDIA RTX 2080 Ti (11 GB); large YAGO/CoDEx runs and distance-decoder spot-checks are therefore server-side experiments. We retain per-run JSONs/logs, seed lists, configs, and code snapshots, and will release an anonymized audit bundle with the final version; all reported metrics use average-rank tie-breaking.

\paragraph{Baselines.}
Baselines use published numbers for classical, neural, and path-based KGC models~\cite{transe,distmult,complex,lacroix2018complex,rotate,compgcn,hake,nbfnet}. They are orientation points, not recipe-matched competitors.

\subsection{Main Results}
\label{sec:main}

\begin{table*}[!htbp]
\centering
\caption{Orientation and selected controlled results under filtered evaluation with average-rank tie-breaking. Published baselines and our controlled runs use different recipes and implementations; the table is not a recipe-matched SOTA comparison.}
\label{tab:main}
\small
\setlength{\tabcolsep}{3pt}
\begin{tabular}{llcccc}
\toprule
Dataset & Method & MRR & Hits@1 & Hits@3 & Hits@10 \\
\midrule
\multirow{5}{*}{FB15k-237}
& RotatE        & $0.338$ & $24.1$ & $37.5$ & $53.3$ \\
& CompGCN       & $0.355$ & $26.4$ & $39.0$ & $53.5$ \\
& KG-ICL-6L     & $0.355$ & $27.0$ & $38.9$ & $52.3$ \\
& NBFNet        & $0.415$ & $32.1$ & $45.4$ & $59.9$ \\
& \ours\ ($L{=}3$, ComplEx) & $0.420{\pm}0.002$ & $32.6{\pm}0.1$ & $46.2{\pm}0.3$ & $60.8{\pm}0.3$ \\
\midrule
\multirow{6}{*}{WN18RR}
& RotatE        & $0.476$ & $42.8$ & $49.2$ & $57.1$ \\
& CompGCN       & $0.479$ & $44.3$ & $49.4$ & $54.6$ \\
& KG-ICL-6L     & $0.442$ & $39.7$ & $46.5$ & $52.4$ \\
& NBFNet        & $0.551$ & $49.7$ & $57.3$ & $66.6$ \\
& \ours\ ($L{=}0$, ComplEx) & $0.485{\pm}0.002$ & $45.6{\pm}0.3$ & $50.1{\pm}0.1$ & $53.5{\pm}0.3$ \\
& \ours\ ($L{=}2$, RotatE) & $0.493{\pm}0.002$ & $45.8{\pm}0.4$ & $50.6{\pm}0.2$ & $56.2{\pm}0.3$ \\
\midrule
\multirow{5}{*}{YAGO3-10}
& ComplEx (in NBFNet)     & $0.587$ & $51.0$ & $63.1$ & $70.0$ \\
& ComplEx-N3 ($d{=}2000$) & $\approx 0.58$ & --- & --- & --- \\
& NBFNet                  & $0.563$ & $48.0$ & $61.0$ & $70.8$ \\
& \ours\ ($L{=}0$, $d{=}128$, 3 seeds) & $0.697{\pm}0.005$ & $63.9{\pm}0.7$ & $73.7{\pm}0.3$ & $79.3{\pm}0.5$ \\
& \ours\ + N3 ($L{=}0$, $d{=}128$) & $0.695{\pm}0.003$ & $63.5{\pm}0.3$ & $72.8{\pm}0.2$ & $79.7{\pm}0.3$ \\
\midrule
\multirow{3}{*}{CoDEx-M}
& ComplEx (Safavi 2020)         & $0.337$ & $26.2$ & --- & $47.6$ \\
& \ours\ ($L{=}0$, $d{=}200$)   & $0.538{\pm}0.000$ & $43.0{\pm}0.1$ & --- & $74.5{\pm}0.2$ \\
& \ours\ + N3 ($L{=}0$, $d{=}200$) & $0.547{\pm}0.003$ & $43.8{\pm}0.4$ & --- & $75.7{\pm}0.3$ \\
\midrule
\multirow{2}{*}{CoDEx-L}
& \ours\ ($L{=}2$, $d{=}200$)   & $0.480{\pm}0.007$ & $38.5{\pm}0.7$ & --- & $66.3{\pm}0.6$ \\
& \ours\ ($L{=}0$, $d{=}200$)   & $0.540{\pm}0.003$ & $44.1{\pm}0.2$ & --- & $72.7{\pm}0.5$ \\
\bottomrule
\end{tabular}
\end{table*}

\paragraph{Five datasets, three verdicts.}
Table~\ref{tab:main} is an orientation table for positioning \ours, not a superiority claim across recipes. On \textbf{YAGO3-10}, pure ComplEx ($L{=}0$, $d{=}128$) reaches MRR $0.6971 \pm 0.0048$ in the new server rerun; adding CompGCN reduces MRR to $0.678 \pm 0.009$. On \textbf{WN18RR}, RotatE reaches $0.493 \pm 0.002$ but still trails NBFNet's $0.551$, keeping path reasoning as a separate lever. On \textbf{FB15k-237}, the encoder helps ($0.406 \to 0.420$ MRR), but the NBFNet gap is within recipe-comparison uncertainty. On \textbf{CoDEx-M/L}, pure ComplEx is the strongest of our structural configurations, while CompGCN hurts most on CoDEx-L.

\begin{figure}[!htbp]
\centering
\includegraphics[width=\columnwidth]{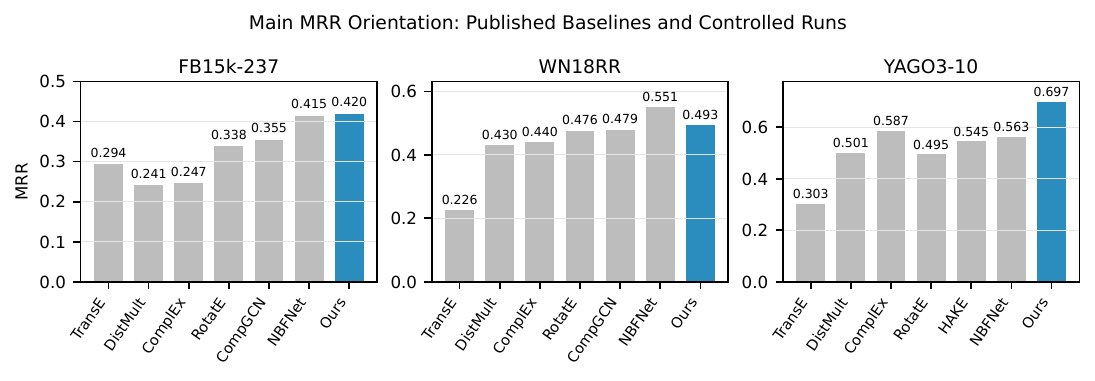}
\caption{Main MRR comparison on the standard transductive benchmarks. The figure is intended as orientation across published baselines and our controlled runs; WN18RR remains favourable to path-based reasoning.}
\label{fig:main-mrr}
\end{figure}

\subsection{The Decoder $\times$ Dataset Diagnostic}
\label{sec:decoder-diagnostic}

We now run the lever-by-lever audit. The central question is not whether decoders are globally more important than encoders, but whether hidden decoder choices can change the interpretation of structural KGC experiments.

\begin{table}[!htb]
\centering
\caption{Diagnostic axes under our fixed recipe (mean MRR; UMLS/Kinship 6 seeds, others 3; std $<\!0.01$ on all cells). \textbf{Dec.\ $\Delta$} = $\mathrm{MRR}_{\mathrm{CX}} - \mathrm{MRR}_{\mathrm{DM}}$; \textbf{Enc.\ $\Delta$} = $\mathrm{MRR}_{L \ge 2} - \mathrm{MRR}_{L=0}$ on ComplEx. Encoder cells are available for the five standard KGs, while UMLS/Kinship are decoder-only small-KG diagnostics.}
\label{tab:decoder-diagnostic}
\small
\setlength{\tabcolsep}{4pt}
\begin{tabular}{lcccc}
\toprule
Dataset & CX MRR & DM MRR & \textbf{Dec.\ $\Delta$} & Enc.\ $\Delta$ \\
\midrule
UMLS        & $\mathbf{0.943}$ & $0.921$ & $+0.022$ & --- \\
Kinship     & $\mathbf{0.853}$ & $0.710$ & $\mathbf{+0.143}$ & --- \\
\midrule
FB15k-237   & $\mathbf{0.421}$ & $0.416$ & $+0.005$ & $+0.014$ \\
WN18RR      & $\mathbf{0.478}$ & $0.466$ & $+0.012$ & $-0.007$ \\
YAGO3-10    & $\mathbf{0.671}$ & $0.665$ & $+0.006$ & $-0.019$ \\
CoDEx-M     & $\mathbf{0.538}$ & $0.528$ & $+0.010$ & $-0.049$ \\
CoDEx-L     & $\mathbf{0.540}$ & $0.536$ & $+0.005$ & $-0.061$ \\
\midrule
\multicolumn{3}{r}{\emph{Shared five-dataset spread:}} & $0.007$ & $0.075$ \\
\multicolumn{3}{r}{\emph{Decoder-only seven-dataset spread:}} & $\mathbf{0.138}$ & --- \\
\bottomrule
\end{tabular}
\end{table}

\paragraph{Shared-grid and small-KG readings.}
On the five standard KGs where both axes are available, decoder $\Delta$ is modest ($+0.005$ to $+0.012$, spread $0.007$), while the CompGCN-style encoder effect varies more ($+0.014$ to $-0.061$, spread $0.075$). This shared-grid result argues against a universal ``decoder dominates encoder'' claim. The decoder-only seven-dataset view is different: adding UMLS and Kinship expands the ComplEx-vs-DistMult range to $0.138$ MRR, driven by Kinship and moderated by UMLS provenance sensitivity. The audit lesson is therefore conditional: decoder choice is a small but consistent reporting factor on standard KGs, and a potentially decisive confound on small KGs.

\begin{figure}[!htbp]
\centering
\includegraphics[width=\columnwidth]{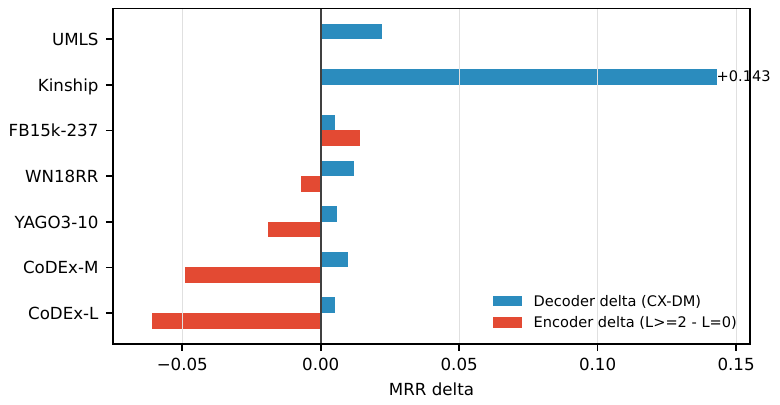}
\caption{Decoder $\Delta$ (ComplEx $-$ DistMult) and encoder $\Delta$ ($L{\geq}2$ vs.\ $L{=}0$ on ComplEx) by dataset. Standard-KG decoder gaps are small, while the decoder-only small-KG rows show why provenance-aware matched decoder checks matter.}
\label{fig:dec-vs-enc}
\end{figure}

\paragraph{The decoder $\Delta$ is small on standard KGs and large on small KGs.}
On the five standard transductive benchmarks (FB15k-237, WN18RR, YAGO3-10, CoDEx-M, CoDEx-L), ComplEx is ahead of DistMult by a modest $+0.005$ to $+0.012$ MRR (Table~\ref{tab:decoder-diagnostic}). On the two small biomedical / kinship benchmarks the gap is much larger, and UMLS changes winner across recipe variants. This motivates the small-KG diagnostic in Sec.~\ref{sec:umls-spotlight}.

\subsection{Small-KG Sensitivity: UMLS and Kinship}
\label{sec:umls-spotlight}

\paragraph{A recipe-sensitive UMLS audit.}
A widely-used biomedical-KG tutorial states that ``DistMult performs poorly on these datasets [UMLS, Kinship] as many relations are antisymmetric in UMLS''~\cite{graph4nlp_kgc}. Our new server rerun under the declared recipe gives UMLS ComplEx MRR $=0.9427 \pm 0.0072$ vs.\ DistMult $=0.9205 \pm 0.0051$ (6 seeds), so ComplEx is ahead by $+0.022$ MRR. However, an earlier run bundle gave DistMult $=0.8705 \pm 0.0109$ vs.\ ComplEx $=0.8344 \pm 0.0020$. We therefore treat UMLS as sensitivity evidence rather than a dataset-level decoder-winner claim. Its role is diagnostic: small-KG decoder conclusions can change across provenance/recipe variants that are often omitted from benchmark tables.

\paragraph{Symmetry is not enough.}
The natural first explanation is that UMLS is symmetric-rich, so DistMult's enforced symmetry acts as a useful bias. Define $\mathrm{sym}(r) = |\{(h,t):(h,r,t),(t,r,h) \in \mathcal{D}_{\text{train}}\}| / |\{(h,t):(h,r,t) \in \mathcal{D}_{\text{train}}\}|$, edge-weighted across relations: UMLS scores $0.133$, Kinship $0.207$ --- UMLS is in fact less symmetric than Kinship. Dataset-level symmetry therefore cannot explain either the earlier UMLS reversal or the larger Kinship ComplEx advantage by itself.

\paragraph{Label smoothing alone is not the sensitivity source.}
In the earlier lock-in pass, we repeated UMLS at $\varepsilon \in \{0, 0.1, 0.3\}$ (3 seeds each, $d{=}200$, $L{=}2$): DistMult won at every $\varepsilon$, with decoder $\Delta$ MRR $= -0.0175$ ($\varepsilon{=}0$), $-0.0268$ ($\varepsilon{=}0.1$), $-0.0446$ ($\varepsilon{=}0.3$). This rules out label smoothing as the sole cause of that pass, but the clean server rerun shows the reversal is not robust enough for a headline claim.

\paragraph{First-cut descriptor: sample size per relation.}
ComplEx doubles DistMult's relation parameter budget via real/imaginary parts, but extra capacity alone is not the whole story. In an earlier lock-in $d \in \{100,200,400\}$ scan, ComplEx is monotone in $d$ on UMLS while DistMult has a $d{=}200$ sweet spot; the clean rerun nevertheless changes the UMLS winner. The most useful descriptor we found is training edges per relation: UMLS has only $113$ e/r and Kinship $342$, far below FB15k-237 ($1148$), CoDEx-M ($3627$), WN18RR ($7894$), and YAGO3-10 ($29\,163$). Synthetic per-relation subsamples of FB15k-237 and WN18RR reproduce a low-e/r DistMult preference in sign, but absolute MRR is low under aggressive subsampling, so e/r should be read as a hypothesis-generating audit descriptor rather than a mechanism proof or winner rule.

\begin{table}[!htbp]
\centering
\caption{Edges per relation as a first-cut audit descriptor. CoDEx-L is excluded from this predictor analysis although its $L{=}0$ decoder spot-check is reported in Table~\ref{tab:decoder-diagnostic}.}
\label{tab:edges-per-rel}
\footnotesize
\setlength{\tabcolsep}{3pt}
\begin{tabular}{lrrrr}
\toprule
Dataset & \#train & \#rel & e/r & Dec.\ $\Delta$ \\
\midrule
UMLS      & 5{,}216 & 46 & $\mathbf{113}$ & $+0.022$ \\
Kinship   & 8{,}544 & 25 & 342 & $\mathbf{+0.143}$ \\
FB15k-237 & 272{,}115 & 237 & 1{,}148 & $+0.005$ \\
CoDEx-M   & 185{,}000 & 51 & 3{,}627 & $+0.010$ \\
WN18RR    & 86{,}835 & 11 & 7{,}894 & $+0.012$ \\
YAGO3-10  & 1{,}079{,}040 & 37 & 29{,}163 & $+0.006$ \\
\bottomrule
\end{tabular}
\end{table}

\begin{figure}[!htbp]
\centering
\includegraphics[width=\columnwidth]{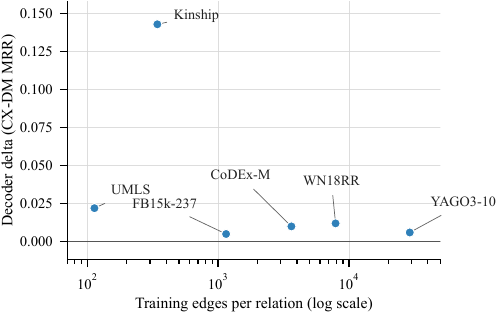}
\caption{Decoder preference as a function of training edges per relation. Low-e/r datasets produce the largest decoder gaps, but UMLS shows that e/r is an audit descriptor rather than a winner rule.}
\label{fig:edges-per-rel}
\end{figure}

\begin{table}[!htbp]
\centering
\caption{Targeted distance-decoder spot-checks ($d{=}200$, $L{=}2$). RotatE/TransE use memory-safe batch/chunk settings on the RTX 2080 Ti, so the table is diagnostic and not recipe-matched to the primary DistMult/ComplEx pair.}
\label{tab:decoder-spotcheck}
\footnotesize
\setlength{\tabcolsep}{3pt}
\begin{tabular}{lcc}
\toprule
Dataset & RotatE & TransE \\
\midrule
UMLS    & $0.8490{\pm}0.0065$ & $0.1005{\pm}0.0087$ \\
Kinship & $0.7266{\pm}0.0091$ & $0.0472{\pm}0.0044$ \\
WN18RR  & $0.4929{\pm}0.0019$ & $0.2016{\pm}0.0045$ \\
\bottomrule
\end{tabular}
\end{table}

\subsection{YAGO3-10: Saturation Under Our Recipe at $d \approx 128$}
\label{sec:yago-saturation}

\paragraph{Setup and scope.}
\cite{lacroix2018complex} reported $\approx\!0.58$ MRR on YAGO3-10 with ComplEx-N3 at $d{=}2000$ (Adagrad, sampled-negatives logistic loss). The original paper introduced N3 regularisation; it did not explicitly assert that capacity \emph{is} the bottleneck. We do not argue against the Lacroix-2018 result itself. Our dimension scan ($L{=}0$ ComplEx, our recipe held fixed; 3 seeds through $d{=}128$ plus a single-seed $d{=}256$ spot-check) shows recipe-conditional near-saturation well below $d{=}2000$:

\begin{table}[!htb]
\centering
\caption{YAGO3-10 capacity scan. $L{=}0$ ComplEx. MRR is near saturation by $d{=}128$; a new $d{=}256$ spot-check gives $+0.008$ at $2\times$ params. The new $d{=}128$ server rerun gives $0.6971 \pm 0.0048$.}
\label{tab:yago-saturation}
\small
\setlength{\tabcolsep}{3pt}
\begin{tabular}{lcccc}
\toprule
$d$ & 64 & 100 & 128 & 256 \\
\midrule
MRR     & $0.671$ & $0.691$ & $0.697$ & $0.705$ \\
$\Delta$ vs prev & --- & $+0.019$ & $+0.006$ & $+0.008$ \\
Params (M) & 7.9 & 12.3 & 15.8 & 31.6 \\
\bottomrule
\end{tabular}
\end{table}

\paragraph{Reading.}
Under our recipe with the ComplEx decoder, the $d{=}64 \to 100$ step gives $+0.019$ MRR, while later increases are smaller: $+0.006$ at $d{=}128$ and $+0.008$ in a single-seed $d{=}256$ spot-check. Thus $d{=}128$ is already near saturation for this configuration, although not a hard optimum. We do not claim this saturation transfers to all recipes, optimisers, or decoders; nor that the Lacroix-2018 number is reproducible \emph{at $d{=}128$} under their original recipe (we did not re-run their setup at our $d$, but doing so would directly test the recipe-vs-capacity confound). Adding N3 to our recipe at $d{=}128$ changes YAGO3-10 MRR by $-0.002$, suggesting N3 and our recipe are largely orthogonal \emph{at this $d$}; we make no claim about N3 at $d{=}2000$ or under other recipes.

\paragraph{Where the YAGO3-10 gain comes from.}
A per-relation audit shows that the aggregate is driven by frequent dense relations rather than by a few tiny relations: \emph{playsFor} and \emph{isAffiliatedTo} alone cover roughly 60\% of the YAGO3-10 test triples and both reach high MRR under the pure ComplEx recipe. Low-MRR relations such as \emph{edited}, \emph{actedIn}, and \emph{influences} are long-tail or require textual/contextual cues, so the result should be read as a structural recipe fit to dense sub-schema rather than a universal KG-completion solution.

\begin{figure}[!htbp]
\centering
\includegraphics[width=\columnwidth]{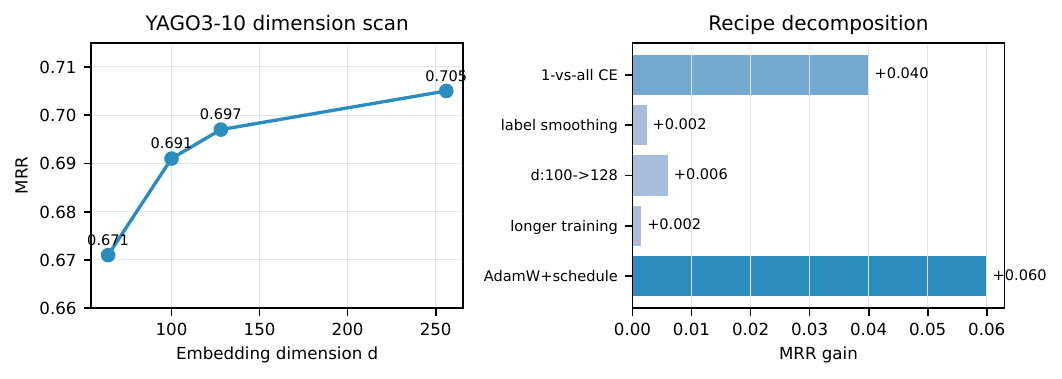}
\caption{YAGO3-10 spotlight: dimension saturation and leave-one-out recipe diagnostics. The difference from older ComplEx settings is associated mainly with the loss and optimiser/learning-rate schedule in this implementation, not simply with increasing $d$.}
\label{fig:yago-spotlight}
\end{figure}

\begin{table}[!htbp]
\centering
\caption{Recipe diagnostics on YAGO3-10. Leave-one-out from our selected configuration; ``original'' denotes the Trouillon-2016 ComplEx setting.}
\label{tab:decomposition}
\footnotesize
\setlength{\tabcolsep}{3pt}
\begin{tabular}{lc}
\toprule
Recipe component (original $\to$ ours) & $\Delta$ MRR \\
\midrule
Logistic loss + 50 sampled neg.\ $\to$ 1-vs-all CE & $+0.040$ \\
$\varepsilon = 0$ $\to$ $\varepsilon = 0.3$ & $+0.0024$ \\
$d = 100$ $\to$ $d = 128$ & $+0.006$ \\
50 epochs $\to$ 150 epochs & $+0.0015$ \\
Adagrad $\to$ AdamW + MultiStepLR & $+0.060$ \\
\midrule
\textbf{Total} & $+0.110$ \\
\bottomrule
\end{tabular}
\end{table}

\subsection{Encoder Depth $\times$ Decoder Choice Interact on WN18RR}
\label{sec:depth-decoder-interaction}

\paragraph{The interaction.}
Most CompGCN-lineage work fixes the decoder (typically ComplEx or DistMult), sweeps the encoder depth $L$, and reports the optimal depth as if it were a property of the dataset. We measure this assumption on WN18RR ($d{=}200$, recipe fixed, 3 seeds per cell):

\begin{table}[!htb]
\centering
\caption{WN18RR encoder-depth $\times$ decoder-choice ($d{=}200$, 3 seeds per cell, mean MRR). Decoders peak at opposite ends ($L{=}0$ ComplEx, $L{=}3$ DistMult) and share an $L{=}1$ dip. For DistMult, $L{=}0$ ties the $L{=}3$ peak within seed std.}
\label{tab:depth-decoder}
\small
\setlength{\tabcolsep}{4pt}
\begin{tabular}{lcccc}
\toprule
Decoder $\backslash$ $L$ & $L{=}0$ & $L{=}1$ & $L{=}2$ & $L{=}3$ \\
\midrule
ComplEx  & $\mathbf{0.485}$ & $0.453$ & $0.481$ & $0.478$ \\
DistMult & $0.473$ & $0.441$ & $0.467$ & $\mathbf{0.474}$ \\
\bottomrule
\end{tabular}
\end{table}

\paragraph{Peaks at opposite ends; shared $L{=}1$ dip.}
Three robust patterns appear: ComplEx peaks at $L{=}0$ ($0.4849$), DistMult at $L{=}3$ ($0.4743$); both decoders share an $L{=}1$ dip; and DistMult's $L{=}3$ peak is only $+0.0012$ MRR above $L{=}0$, within seed std. A fixed-decoder ablation would conclude ``encoder hurts'' or ``encoder neutral'', while a fixed-encoder ablation would conclude ``ComplEx $>$ DistMult''. Neither captures the $4{\times}2$ interaction.

\paragraph{Implication.}
Encoder utility is dataset- and decoder-shaped, not uniformly low. On FB15k-237, the encoder helps ComplEx by $+0.014$ MRR; on WN18RR, it is neutral or harmful for ComplEx. Small KGs differ again: $L{=}0$ collapses on UMLS/Kinship, so the WN18RR observation does not generalise to all sparse settings.

\subsection{Training Recipe Ablation}
\label{sec:ablation}

To identify which training choices drive our results, we ablate six ingredients of the recipe on WN18RR (3 seeds) and FB15k-237 (single seed, 3-seed for headline rows), starting from the main configuration for each dataset.

\begin{table}[!htb]
\centering
\caption{Training-recipe ablation on WN18RR ($d{=}200$, $L{=}2$, 3 seeds; mean MRR with $H@k$ in \%). Label smoothing is the most impactful single ingredient.}
\label{tab:ablation-wn18rr}
\footnotesize
\setlength{\tabcolsep}{3pt}
\begin{tabular}{lcccc}
\toprule
Configuration & MRR & H@1 & H@3 & H@10 \\
\midrule
Baseline (d=200, L=2, LS=0.3, DM) & $0.467$ & $43.6$ & $48.0$ & $52.5$ \\
\midrule
$+$ deeper GCN (L=3)            & $\mathbf{0.474}$ & $44.3$ & $48.9$ & $\mathbf{53.1}$ \\
$+$ ComplEx decoder             & $\mathbf{0.478}$ & $\mathbf{45.1}$ & $\mathbf{49.1}$ & $52.8$ \\
$-$ smaller dim (d=128)         & $0.466$ & $43.6$ & $47.8$ & $52.0$ \\
$-$ shallower (L=1)             & $0.441$ & $41.7$ & $45.1$ & $48.7$ \\
$-$ weaker smoothing (LS=0.1)   & $0.457$ & $43.0$ & $46.7$ & $50.6$ \\
$-$ no smoothing (LS=0.0)       & $0.427$ & $40.7$ & $43.4$ & $46.1$ \\
\bottomrule
\end{tabular}
\end{table}

\paragraph{Label smoothing is dominant.}
Removing label smoothing ($\varepsilon = 0$) drops MRR by $-0.041$ on WN18RR (Table~\ref{tab:ablation-wn18rr}) and $-0.017$ on FB15k-237; reducing to $\varepsilon = 0.1$ already costs $-0.010$/$-0.020$. This is larger than any architectural change we tested.

\begin{figure}[!htbp]
\centering
\includegraphics[width=\columnwidth]{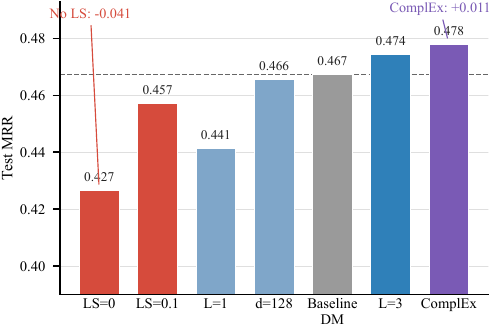}
\caption{WN18RR recipe ablation. Label smoothing is the largest single ingredient in this recipe, while depth and decoder changes are smaller and interact with each other.}
\label{fig:ablation-wn18rr}
\end{figure}

\paragraph{Shallow GCN, decoder picks up the slack.}
A single-layer GCN loses $-0.029$ MRR on WN18RR; $L{=}2 \to L{=}3$ gives $+0.004$ on WN18RR, $+0.010$ on FB15k-237, but reverses sign on YAGO3-10 ($-0.022$). Replacing DistMult with ComplEx yields $+0.011$ on WN18RR, $+0.003$ on FB15k-237, and $+0.006$--$0.011$ across $L$ on YAGO3-10. The cross-dataset decoder pattern generalises to all five transductive benchmarks (Sec.~\ref{sec:decoder-diagnostic}).

\subsection{YAGO3-10 Recipe Diagnostics plus Sensitivity / Efficiency}
\label{sec:decomposition}\label{sec:sensitivity}

Our $L{=}0$ ComplEx on YAGO3-10 ($0.6971$) is $+0.110$ MRR relative to the Trouillon-2016 ComplEx number reported in the NBFNet comparison table~\cite{nbfnet}. A 3-seed leave-one-out diagnostic associates the difference with the loss change ($+0.040$), smoothing/training length/$d$ ($+0.010$), and optimiser/initialisation residual ($+0.060$). We read this as implementation-level sensitivity rather than a causal proof that any single ingredient dominates. N3 is largely orthogonal at saturated $d$. The resulting configuration is also computationally light after graph encoding: on one RTX 2080 Ti, $L{=}0$ ComplEx scores about $112$K YAGO3-10 queries/s at $d{=}128$, while $L{=}2$ DistMult at $d{=}64$ scores about $297$K queries/s after a $25.9$ ms graph-encoding pass.

\section{Discussion}
\label{sec:discussion}

\paragraph{The WN18RR gap to NBFNet.}
\label{sec:wn18rr-gap}
We do \emph{not} claim superiority over path-based models: on WN18RR, NBFNet's $0.551$ stays $5.8$ MRR above our RotatE spot-check ($0.493$) and $6.6$ above our selected ComplEx configuration ($0.485$). The gap is consistent with path propagation having a genuine advantage on tree-like hypernym chains. Path-based reasoning is a separate lever, orthogonal to our decoder-vs-CompGCN-encoder audit.

\section{Conclusion}

Prior controlled audits cover recipe~\cite{ruffinelli2020olderboys} and CompGCN-style encoder~\cite{zhang2022rethinking} axes; this paper adds the decoder axis through \ours. The actionable outcome is a compact audit checklist: report a matched DistMult-vs-ComplEx row, log small-KG recipe and provenance details, inspect edges per relation and symmetry as descriptors, and avoid reporting encoder depth without the decoder dimension. On standard KGs, decoder gaps are modest but consistent under our recipe; on small KGs, they can become decisive and provenance-sensitive. Future encoder ablations should report the full $\{\mathrm{decoder}\} \times \{L\}$ grid rather than a row-wise minimum.

\section*{Limitations}
\label{sec:limitations}

Our conclusions are limited to structural transductive KGC with the ComplEx/DistMult decoder family, CompGCN-style encoders, and one fixed recipe; they do not extend to inductive, text-augmented, LLM-based, OGB-scale, or path-based KGC. A fixed recipe isolates decoder and encoder effects but does not identify each model's independently tuned optimum, and recipe itself is a large lever. Table~\ref{tab:main} mixes published baselines and our recipe-controlled runs, so small gaps should be read cautiously. The decoder and encoder $\Delta$ columns are not fully symmetric: UMLS/Kinship are decoder-only small-KG diagnostics, seed budgets and depths differ, and YAGO3-10 dimensionality differs across cells. The UMLS mechanism remains a heuristic: edge-weighted symmetry is coarse, UMLS/Kinship are small benchmarks, e/r subsampling can reduce absolute MRR, and UMLS changes winner between an earlier run bundle and the clean server rerun. Finally, throughput is scoring throughput after graph encoding, not full end-to-end training or inference wall-clock.

\section*{Ethics Statement}

Improved KG completion benefits downstream applications (search, recommendation, biomedical KG construction) but can amplify biases in the source KG. Our recipe reduces GPU-hours per trained model and brings reproduction within reach of single-11-GB-GPU groups. We encourage per-relation error analysis and fairness auditing before deployment. All datasets used are public.

\bibliographystyle{splncs04}
\bibliography{references}

@inproceedings{transe,
  title={Translating embeddings for modeling multi-relational data},
  author={Bordes, Antoine and Usunier, Nicolas and Garcia-Dur{\'a}n, Alberto and Weston, Jason and Yakhnenko, Oksana},
  booktitle={Advances in Neural Information Processing Systems},
  year={2013}
}

@inproceedings{distmult,
  title={Embedding entities and relations for learning and inference in knowledge bases},
  author={Yang, Bishan and Yih, Wen-tau and He, Xiaodong and Gao, Jianfeng and Deng, Li},
  booktitle={International Conference on Learning Representations},
  year={2015}
}

@inproceedings{complex,
  title={Complex embeddings for simple link prediction},
  author={Trouillon, Th{\'e}o and Welbl, Johannes and Riedel, Sebastian and Gaussier, {\'E}ric and Bouchard, Guillaume},
  booktitle={International Conference on Machine Learning},
  year={2016}
}

@inproceedings{rotate,
  title={{RotatE}: Knowledge graph embedding by relational rotation in complex space},
  author={Sun, Zhiqing and Deng, Zhi-Hong and Nie, Jian-Yun and Tang, Jian},
  booktitle={International Conference on Learning Representations},
  year={2019}
}

@inproceedings{rgcn,
  title={Modeling relational data with graph convolutional networks},
  author={Schlichtkrull, Michael and Kipf, Thomas N and Bloem, Peter and van den Berg, Rianne and Titov, Ivan and Welling, Max},
  booktitle={European Semantic Web Conference},
  year={2018}
}

@inproceedings{compgcn,
  title={Composition-based multi-relational graph convolutional networks},
  author={Vashishth, Shikhar and Sanyal, Soumya and Nitin, Vikram and Talukdar, Partha},
  booktitle={International Conference on Learning Representations},
  year={2020}
}

@inproceedings{nbfnet,
  title={Neural {Bellman-Ford} networks: A general graph neural network framework for link prediction},
  author={Zhu, Zhaocheng and Zhang, Zuobai and Xhonneux, Louis-Pascal and Tang, Jian},
  booktitle={Advances in Neural Information Processing Systems},
  year={2021}
}

@inproceedings{astarnet,
  title={{A*Net}: A scalable path-based reasoning approach for knowledge graphs},
  author={Zhu, Zhaocheng and Yuan, Xinyu and Galkin, Mikhail and Xhonneux, Sophie and Zhang, Ming and Gazeau, Maxime and Tang, Jian},
  booktitle={Advances in Neural Information Processing Systems},
  year={2023}
}

@inproceedings{redgnn,
  title={Knowledge graph reasoning with relational digraph},
  author={Zhang, Yongqi and Yao, Quanming},
  booktitle={The Web Conference},
  year={2022}
}

@inproceedings{hake,
  title={Learning hierarchy-aware knowledge graph embeddings for link prediction},
  author={Zhang, Zhanqiu and Cai, Jianyu and Zhang, Yongdong and Wang, Jie},
  booktitle={AAAI Conference on Artificial Intelligence},
  year={2020}
}

@inproceedings{fb15k237,
  title={Representing text for joint embedding of text and knowledge bases},
  author={Toutanova, Kristina and Chen, Danqi and Pantel, Patrick and Poon, Hoifung and Choudhury, Pallavi and Gamon, Michael},
  booktitle={Conference on Empirical Methods in Natural Language Processing},
  year={2015}
}

@inproceedings{wn18rr,
  title={Convolutional 2D knowledge graph embeddings},
  author={Dettmers, Tim and Minervini, Pasquale and Stenetorp, Pontus and Riedel, Sebastian},
  booktitle={AAAI Conference on Artificial Intelligence},
  year={2018}
}

@inproceedings{yago310,
  title={{YAGO3}: A knowledge base from multilingual {Wikipedias}},
  author={Mahdisoltani, Farzaneh and Biega, Joanna and Suchanek, Fabian M},
  booktitle={Conference on Innovative Data Systems Research (CIDR)},
  year={2015}
}

@inproceedings{safavi2020codex,
  title={{CoDEx}: A Comprehensive Knowledge Graph Completion Benchmark},
  author={Safavi, Tara and Koutra, Danai},
  booktitle={Conference on Empirical Methods in Natural Language Processing (EMNLP)},
  year={2020}
}

@inproceedings{balazevic2019tucker,
  title={{TuckER}: Tensor Factorization for Knowledge Graph Completion},
  author={Bala{\v{z}}evi{\'c}, Ivana and Allen, Carl and Hospedales, Timothy M},
  booktitle={Conference on Empirical Methods in Natural Language Processing (EMNLP)},
  year={2019}
}

@inproceedings{chen2021hitter,
  title={{HittER}: Hierarchical Transformers for Knowledge Graph Embeddings},
  author={Chen, Sanxing and Liu, Xiaodong and Gao, Jianfeng and Jiao, Jian and Zhang, Ruofei and Ji, Yangfeng},
  booktitle={Conference on Empirical Methods in Natural Language Processing (EMNLP)},
  year={2021}
}

@inproceedings{ruffinelli2020olderboys,
  title={You {CAN} Teach an Old Dog New Tricks! On Training Knowledge Graph Embeddings},
  author={Ruffinelli, Daniel and Broscheit, Samuel and Gemulla, Rainer},
  booktitle={International Conference on Learning Representations (ICLR)},
  year={2020}
}

@inproceedings{lacroix2018complex,
  title={Canonical Tensor Decomposition for Knowledge Base Completion},
  author={Lacroix, Timoth{\'e}e and Usunier, Nicolas and Obozinski, Guillaume},
  booktitle={International Conference on Machine Learning (ICML)},
  pages={2863--2872},
  year={2018}
}

@inproceedings{ali2021hparam,
  title={Bringing Light into the Dark: A Large-scale Evaluation of Knowledge Graph Embedding Models Under a Unified Framework},
  author={Ali, Mehdi and Berrendorf, Max and Hoyt, Charles Tapley and Vermue, Laurent and Galkin, Mikhail and Sharifzadeh, Sahand and Fischer, Asja and Tresp, Volker and Lehmann, Jens},
  booktitle={IEEE Transactions on Pattern Analysis and Machine Intelligence (TPAMI)},
  year={2021}
}

@inproceedings{zhang2022rethinking,
  title={Rethinking Graph Convolutional Networks in Knowledge Graph Completion},
  author={Zhang, Zhanqiu and Wang, Jie and Ye, Jieping and Wu, Feng},
  booktitle={Proceedings of the ACM Web Conference (WWW)},
  year={2022}
}

@inproceedings{manabe2018datadependent,
  title={Data-Dependent Learning of Symmetric/Antisymmetric Relations for Knowledge Base Completion},
  author={Manabe, Hitoshi and Hayashi, Katsuhiko and Shimbo, Masashi},
  booktitle={AAAI Conference on Artificial Intelligence},
  year={2018}
}

@inproceedings{akrami2020realistic,
  title={Realistic Re-evaluation of Knowledge Graph Completion Methods: An Experimental Study},
  author={Akrami, Farahnaz and Saeef, Mohammed Samiul and Zhang, Qingheng and Hu, Wei and Li, Chengkai},
  booktitle={Proceedings of the ACM SIGMOD International Conference on Management of Data},
  year={2020}
}

@inproceedings{sun2020reevaluation,
  title={A Re-evaluation of Knowledge Graph Completion Methods},
  author={Sun, Zhiqing and Vashishth, Shikhar and Sanyal, Soumya and Talukdar, Partha and Yang, Yiming},
  booktitle={Annual Meeting of the Association for Computational Linguistics (ACL)},
  year={2020}
}

@inproceedings{kazemi2018simple,
  title={{SimplE} embedding for link prediction in knowledge graphs},
  author={Kazemi, Seyed Mehran and Poole, David},
  booktitle={Advances in Neural Information Processing Systems},
  year={2018}
}

@article{trouillon2017complex,
  title={Knowledge Graph Completion via Complex Tensor Factorization},
  author={Trouillon, Th{\'e}o and Dance, Christopher R. and Welbl, Johannes and Riedel, Sebastian and Gaussier, {\'E}ric and Bouchard, Guillaume},
  journal={Journal of Machine Learning Research},
  volume={18},
  pages={1--38},
  year={2017}
}

@misc{graph4nlp_kgc,
  title={Knowledge Graph Completion --- {Graph4NLP} v0.4.1 documentation},
  author={{{Graph4NLP} Authors}},
  howpublished={\url{https://graph4ai.github.io/graph4nlp/guide/classification/kgcompletion.html}},
  year={2021},
  note={Accessed 2026-05-04}
}


\end{document}